\documentclass[conference]{IEEEtran}
\usepackage{graphicx}
\usepackage{latexsym}
\usepackage{times}
\usepackage{soul}
\usepackage{url}
\usepackage[hidelinks]{hyperref}
\usepackage[utf8]{inputenc}
\usepackage[small]{caption}
\usepackage{graphicx}
\usepackage{amsmath}
\usepackage{amsthm}
\usepackage{booktabs}
\usepackage{algorithm}
\usepackage{algorithmic}
\usepackage{times}
\usepackage{soul}
\usepackage{url}
\usepackage[hidelinks]{hyperref}
\usepackage[utf8]{inputenc}
\usepackage{graphicx}
\usepackage{amsmath}
\usepackage{amssymb}
\usepackage{booktabs}
\usepackage{algorithm}
\usepackage{algorithmic}
\usepackage[switch]{lineno}
\usepackage{color}
\usepackage{array}
\usepackage{booktabs}
\usepackage{multirow}
\usepackage{bm}
\usepackage{mathrsfs}
\usepackage{enumitem}
\usepackage{amsthm}
\usepackage[misc,geometry]{ifsym}
\makeatletter
\renewcommand\@makefnmark{\hbox{\@textsuperscript{*}}}
\makeatother

\ifCLASSINFOpdf
\else
\fi
\begin{document}
%
\title{Explanation based Bias Decoupling Regularization for Natural Language Inference}

\author{\IEEEauthorblockN{Jianxiang Zang, Hui Liu\textsuperscript{\Letter}}
\IEEEauthorblockA{School of Statistics and Information Science\\
Shanghai University of International Business and Economics\\
Email: \{21349110, liuh\}@suibe.edu.cn}}


%


\maketitle

\begin{abstract}
The robustness of Transformer-based Natural Language Inference encoders is frequently compromised as they tend to rely more on dataset biases than on the intended task-relevant features.
Recent studies have attempted to mitigate this by reducing the weight of biased samples during the training process. However, these debiasing methods primarily focus on identifying which samples are biased without explicitly determining the biased components within each case. This limitation restricts those methods' capability in out-of-distribution inference. To address this issue, we aim to train models to adopt the logic humans use in explaining causality. We propose a simple, comprehensive, and interpretable method: \textbf{E}xplanation based \textbf{B}ias \textbf{D}ecoupling \textbf{Reg}ularization (\textbf{EBD-Reg}). EBD-Reg employs human explanations as criteria, guiding the encoder to establish a tripartite parallel supervision of Distinguishing, Decoupling and Aligning. This method enables encoders to identify and focus on keywords that represent the task-relevant features during inference, while discarding the residual elements acting as biases. Empirical evidence underscores that EBD-Reg effectively guides various Transformer-based encoders to decouple biases through a human-centric lens, significantly surpassing other methods in terms of out-of-distribution inference capabilities.
\end{abstract}


%
\IEEEpeerreviewmaketitle

\section{Introduction}

Transformer-based encoders have shown remarkable capabilities in Natural Language Inference (NLI)~\cite{devlin2019bert,liu2019roberta,lan2019albert,zang2023improving}. However, these models tend to leverage dataset biases rather than the intended task-relevant features. For instance, implication models trained on the MNLI dataset~\cite{bowman2015large} often make predictions based solely on the presence of specific key terms~\cite{gururangan2018annotation} or whether pairs of sentences share the same words~\cite{mccoy2019right}. Current debiasing methods rely on alternate learning~\cite{schuster2019towards,clark2019don,utama2020mind}, which suppress the model's focus on biased samples, thereby challenging it to learn from harder samples. However, these debiasing methods aim to teach models which samples are biased, but they do not specify which parts within an individual sample are biased. Merely identifying biased individual samples limits the model's ability to infer out-of-distribution, as the attributes of individual samples are highly correlated with their respective distributions. Therefore, we aim to train the model to distinguish the biased parts (words) in each sample.

We observe that humans excel at identifying the crux when explaining causal relationships, often focusing on the main contradictions in phenomena. These primary contradictions usually embody elements that differentiate between cause and effect. Inspired by this human aptitude, we thoroughly analyze the inherent connection between human explanations and biases at word level in Section~\ref{sec.3.1}, summarizing criteria for distinguishing keywords and biases from a human perspective. Following this criterion, we introduce \textbf{E}xplanation based \textbf{B}ias \textbf{D}ecoupling \textbf{Reg}ularization (\textbf{EBD-Reg}) to guide models in learning the logic humans use when explaining issues. This approach enables the model to focus on keywords that represent task-relevant features during inference, while discarding elements that act as biases. Leveraging human explanations from the eSNLI dataset~\cite{camburu2018snli}, EBD-Reg establishes a triple parallel supervision training objective: (a) Distinguishing, to identify keywords and biases; (b) Decoupling, to encourage the model to focus on keywords while suppressing attention to biases; (c) Aligning, employing a divide-and-conquer strategy that align the joint predictive distribution of keyword and bias inference with that of the main inference, expecting the model to make more interpretable final predictions based on the condition of decoupling biases from keywords. Furthermore, we introduce \textbf{A}daptive \textbf{T}oken-level \textbf{A}ttention (\textbf{ATA}) to reinforce both main and sub-inferences. Extensive empirical evidence has shown that EBD-Reg can be easily integrated with various Transformer-based encoders, significantly surpassing other debiasing methods in enhancing out-of-distribution inference performance.

Our main contributions resides in the following aspects: 
(1) While traditional NLI debiasing methods teach models "which samples are biased", our approach, rooted in human explanation, aims to instruct models "which parts of a sample are biased".
(2) We conducted a pioneering analysis of the intrinsic connection between human explanations and biases, proposing a method for distinguishing between keywords and biases.
(3) We propose a simple, comprehensive, and interpretable debiasing method, Explanation based Bias Decoupling Regularization (EBD-Reg), which significantly surpasses other debiasing methods in out-of-distribution inference performance.

\section{Related Works}

Due to Transformer-based encoders consistently handling content across various levels of inference granularity within data, models tend to leverage dataset biases rather than intended task-relevant features, leading researchers to propose a series of debiasing methods.
Enhancing the training corpus stands as the predominant method for bias correction, including the autogeneration and identification of adversarial examples~\cite{minervini2018adversarially}, the augmentation of standard training sets with syntactically-rich examples~\cite{min2020syntactic}, fine-tuning on a minority of examples, the introduction of counterfactual instances to pin down the gradient of a model's decision function~\cite{teney2020learning}, and the application of Variational Information Bottleneck (VIB) to suppress irrelevant features within the corpus~\cite{belinkov2020variational}.With the widespread adoption of knowledge distillation, debiasing algorithms utilizing alternative learning approaches have gained increasing popularity.~\cite{clark2019don,utama2020mind,utama2020towards,he2019unlearn,liu2020empirical,sanh2020learning,schuster2019towards}. Such methods can be formalized into a two-stage framework: the first entails the training of a model that is exclusively biased, either automatically~\cite{utama2020mind,sanh2020learning,ghaddar2021end} or utilizing prior knowledge regarding biases~\cite{clark2019don,he2019unlearn,belinkov2019adversarial}. In the second stage, the biased model's output is harnessed to re-calibrate the loss function of the unbiased model. Other debiasing methods include predicting the probability of a premise given a hypothesis and NLI label, countering biases stemming from model disregard of the premise~\cite{belinkov2019don}, employing multi-task learning to bolster model generalization~\cite{tu2020empirical}, and deploying a combination of distant supervision to manage inference information of various granularities in the corpus~\cite{zou2022divide}. 

However, these debiasing methods aim to teach models which samples are biased, but they do not specify which parts within an individual sample are biased. Merely identifying biased individual samples limits the model's ability to infer out-of-distribution, as the attributes of individual samples are highly correlated with their respective distributions.

\section{Revisiting Natural Language Inference}


Natural Language Inference is to develop a classifier $\xi$ that can compute the conditional probabilities $P(y|\bm{s}^{a,b})$ to predict the relationships between the output sentence pairs. Here, $y\in Y$ represents the semantic relationships between the premise and hypothesis, including \{entailed, neutral, contradicted\}. $\bm{s}^{a,b}=\{\text{[CLS]};\bm{s}^a;\text{[SEP]};\bm{s}^b\}$ represents the concatenation of premise $\bm{s}^a=\{s^a_i\}^{l_a}_{i=1}$ and hypothesis $\bm{s}^b=\{s^b_j\}^{l_b}_{j=1}$ using separator tokens $\text{[CLS]}$ and $\text{[SEP]}$. 
For predicting the main objective $P(y|\bm{s}^{a,b})$, we feed $\bm{s}^{a,b}$ into a Transformer-based encoder and obtain global semantic representations $\{\bm{v}_i\}_{i=1}^{l_{a,b}}$, and $\bm{v}_{1}$ denotes the representation of first token $\text{[CLS]}$. $l_a$, $l_b$, and $l_{a,b}$ represent the sequence lengths of the premise, the hypothesis, and the premise-hypothesis pair, respectively.

Inspired by~\cite{rei2018jointly,stacey2022supervising}, we advocate against employing $\bm{v}_{1}$ as the classification head. Instead, we introduce Adaptive Token-level Attention (ATA) for subsequent predictions. Specifically, we obtain token-level predictions $\bm{\lambda}_i$ by passing the global semantic representation $\bm{v}_i$ through two non-linear layers, and then normalize it to serve as adaptive weights $\{\widetilde{\bm{\lambda}}\}_{i=1}^{l_{a,b}}$. In this context, all the $\bm{W}$s mentioned in this paper refer to weight matrices with bias terms included.

\begin{equation}
\begin{gathered}
\bm{\lambda}_i={\rm sigmoid}({\bm{W}_2}\cdot ({\rm tanh}({\bm{W}_1} \cdot \bm{v}_{i})))\\
\widetilde{\bm{\lambda}}_i=\frac{\bm{\lambda}_i}{\sum_{k=1}^{l_{a,b}}\bm{\lambda}_k}
\end{gathered}
\end{equation}

Finally, we perform element-wise multiplication of these weights with the global semantic representation, and use the result after average pooling as the classification head. We then make predictions, and calculate the loss for fine-tuning as formulated in Equation~\ref{eq.main}.


\begin{equation}
\begin{gathered}
     P(y|\bm{s}^{a,b})={\rm softmax}(\bm{W}^{\text{pre}} \cdot (\frac{1}{l_{a,b}}\sum_{i=1}^{l_{a,b}}\bm{\lambda}_i\bm{v}_{i}))\\
    \mathcal{L}_{\rm main}=-{\rm log}P(y|\bm{s}^{a,b})~\label{eq.main}   
\end{gathered}
\end{equation}

\begin{figure*}[t]
\centering
\includegraphics[width=0.80\textwidth]{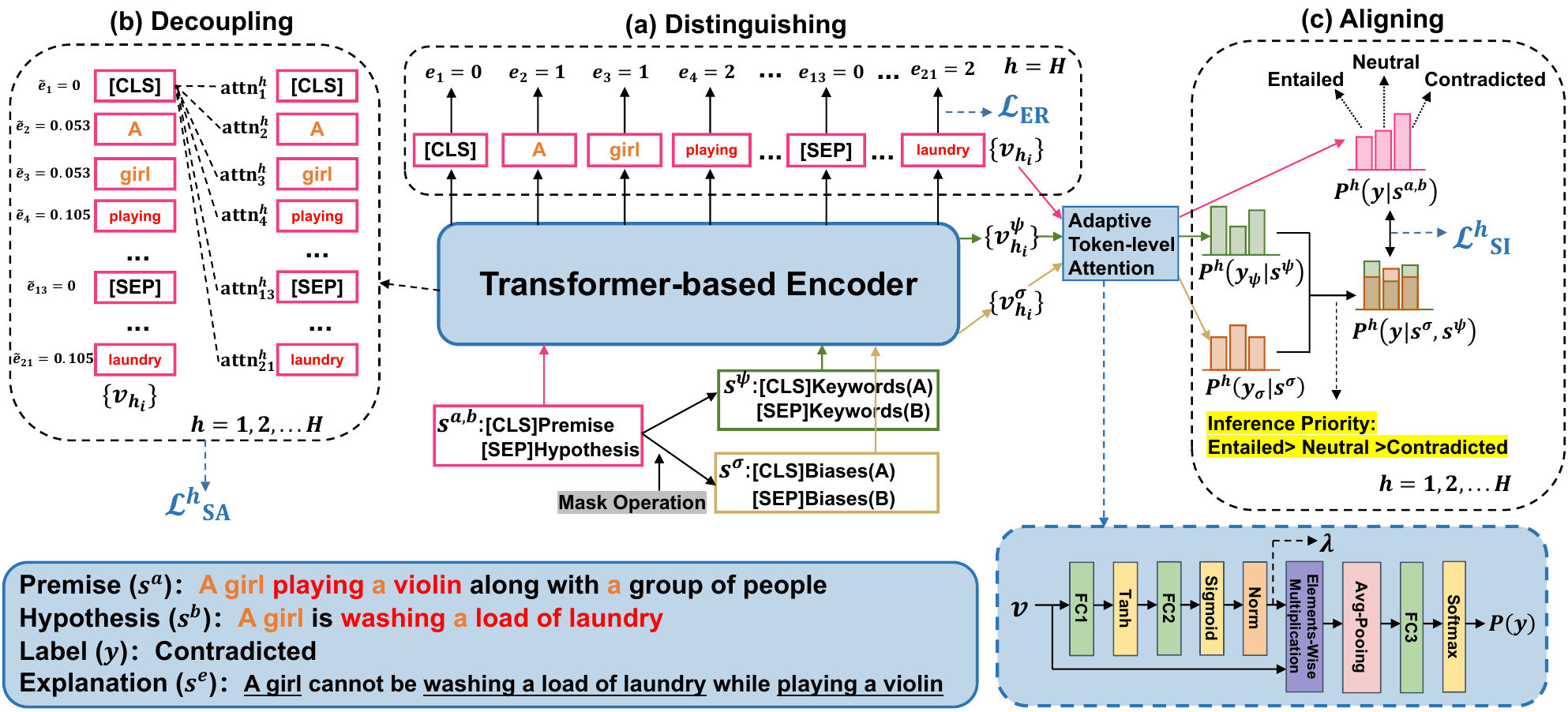}
\caption{Overview of EBD-Reg, which includes 3 targets: (a) Distinguishing; (b) Decoupling; (c) Aligning. 
}\label{fig.decoulper}
\end{figure*}

\begin{figure}[t]
\centering
\includegraphics[width=1\linewidth]{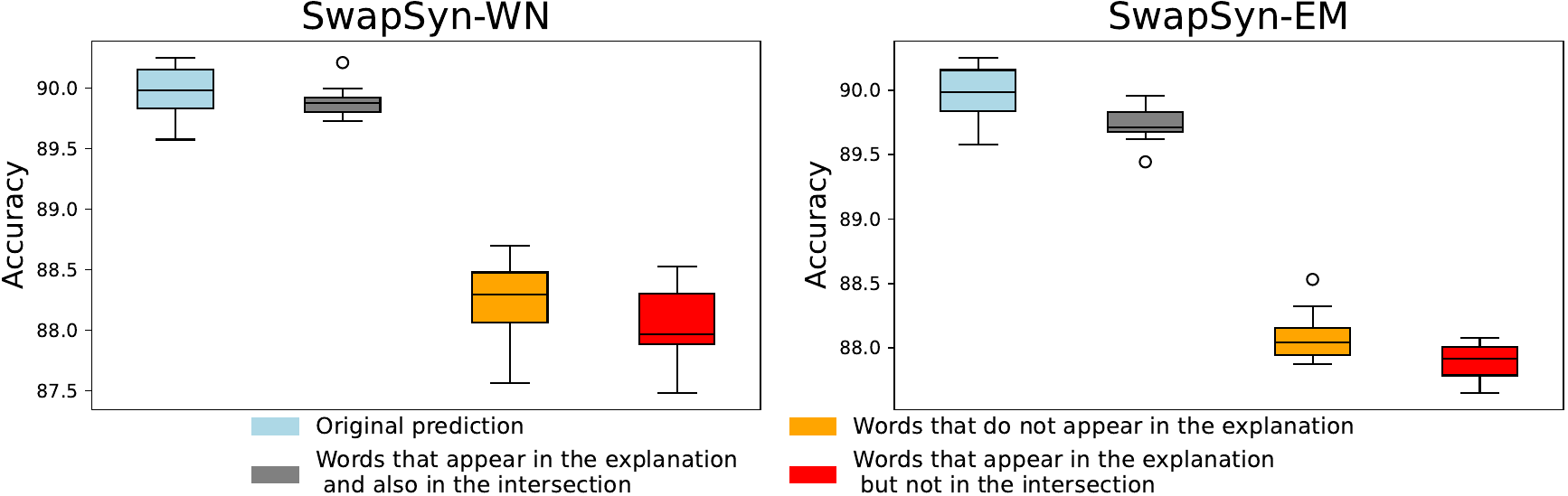}
\caption{
Evaluation results for the eSNLI test set. SwapSyn-WN: Replaces words with synonyms provided by WordNet; SwapSyn-EM: Uses GloVe embeddings to replace common words with synonyms.}\label{fig.Swap}
\end{figure}

\section{Our Method}




Transformer-based encoders, when used for text inference, typically process each word through a uniform comparative approach. However, each sentence can usually be broken down into content with varying levels of matching granularity~\cite{bommasani2021opportunities,su2021keep}. We introduce a task-specific assumption in Natural Language Inference, positing that each sentence can be decomposed into keywords and biases. Here, keywords represent features related to the intended task, while the remaining biased elements could mislead the model's judgment. Traditional debiasing methods focus on identifying which samples are biased, but our approach aims for the model to pinpoint the biased components within.

\subsection{Inspired by Human Explanation}\label{sec.3.1}

The eSNLI dataset~\cite{camburu2018snli} is an expanded version of SNLI~\cite{bowman2015large}, designed to provide explanations for each inference example. The explanations are created by asking annotators from Amazon Mechanical Turk to explain why each hypothesis and premise are assigned their given labels.
 As illustrated in Figure~\ref{fig.decoulper}, for the given premise: "A girl playing a violin along with a group of people" and the hypothesis: "A girl is washing a load of laundry", The explanation for their "contradicted" relationship is: "A girl cannot be washing a load of laundry while playing a violin".

 We observe that humans excel at identifying the crux when explaining causal relationships, often focusing on the main contradictions in phenomena. These primary contradictions usually embody elements that differentiate between cause and effect. Inspired by this, we investigate the relationship between the premise-hypothesis words with different level of inference granularity and explanations. We categorize words into three classes: (1) those not appearing in the explanation (e.g., "is","along","with","group","of","people"), (2) those appearing in both the explanation and the intersection of the premise and hypothesis (e.g., "a","girl"), (3) those appearing in the explanation but not in the intersection (e.g., "playing," "violin," "washing," "laundry"). In the eSNLI test set, we focus on three categories of words, applying transformations using SwapSyn-WN~\cite{miller1995wordnet} and SwapSyn-EM~\cite{pennington2014glove}, with each transformation is performed 8 times. The accuracy of these transformed texts is then assessed using the fine-tuned BERT-SNLI. The Figure~\ref{fig.Swap} illustrates that replacements of words appearing in explanations lead to a noticeable accuracy drop, with further reductions for words not intersecting in the premise and hypothesis, whereas replacements of words not appearing in explanations yield more stable predictions. Based on this experiment, the following criteria for division can be established: words not in the explanation can be considered as biases, potentially misleading the model. Words appearing in explanations should be treated as keywords, demanding heightened attention from the model, especially those not intersecting in the premise and hypothesis, and should be assigned higher importance weights. Based on this conclusion, we explore how EBD-Reg, drawing from human explanations in the eSNLI dataset, constructs a tripartite parallel supervision of Distinguishing, Decoupling and Aligning to achieve the identification and decoupling of biases.

\subsection{Explanation based Bias Decoupling}

\subsubsection{Distinguishing}

We consider that the model must learn to distinguish between keywords and biases in the first step. This understanding is fundamental to the process of bias decoupling. Therefore, we introduce supervision for \textbf{E}ntity \textbf{R}ecognition ($\mathcal{L}_{\rm ER}$) to facilitate this learning. Based on the explanations, given  \(\bm{s}^e=\{s^e_i\}_{i=1}^{l_{e}}\), we create labels \(\bm{e}=\{e_i\}^{l_{a,b}}_{i=1}\) for each observation in \(\bm{s}^{a,b}\), where \(l_e\) is the length of explanation squence. Specifically, based on the aforementioned conclusions in Section~\ref{sec.3.1}, words in \(\bm{s}^{a,b}\) that do not appear in the explanation are assigned a label of 0. As keywords, words that are present in the explanation but not in both \(\bm{s}^a\) and \(\bm{s}^b\) are assigned a label of 2. Finally, words that appear in both \(\bm{s}^a\) and \(\bm{s}^b\) are tagged with a label of 1. 

\begin{equation}
e_i=
\begin{cases}
2,\quad s^{a,b}_i \in \bm{s}^e,\notin \bm{s}^a \cap \bm{s}^b\\
1,\quad s^{a,b}_i \in \bm{s}^e,\in \bm{s}^a \cap \bm{s}^b\\
0,\quad s^{a,b}_i \notin \bm{s}^e\\
\end{cases}
\end{equation}

After obtaining the labels, we introduce a training objective aimed at compelling the encoder to effectively differentiate between keywords and biases.
Specifically, we use the global semantic representation $\bm{v}_i$ and define the loss as formulated in Equation~\ref{eq.er}. Wherein, $P(e_i|\bm{v}_i)$ is the probability in the predicted probability distribution for the $i^{th}$ token, corresponding to the correct label $e_i$.

\begin{equation}
\begin{gathered}
    P(e_i|\bm{v}_i)={\rm softmax}(\bm{W}^{\text{pre}} \cdot \bm{v}_i)\\
    \mathcal{L}_{\rm ER}=-\frac{1}{l_{a,b}}\sum_{i=1}^{l_{a,b}}{\rm log}P(e_i|\bm{v}_i)~\label{eq.er}
\end{gathered}
\end{equation}

\begin{table*}[h]
\centering
\setlength{\tabcolsep}{6pt} 
\renewcommand{\arraystretch}{1} 
\begin{tabular}{llllllllll}
\hline
Model                    & $ \rm SNLI_{test}$ & $\rm SNLI_{hard}$ & $\rm MNLI_{m}$ & $\rm MNLI_{mm}$ & $\rm RTE_{dev}$ & $\rm SciTail_{dev}$ & $\rm QNLI_{dev}$ & ANLI  & HANS  \\ \hline
BERT-eSNLI                & 90.08       & 79.36       & 72.51     & 72.41     & 69.89     & 75.21   & 54.89      & 31.72 & 56.83 \\
+Reweight~\cite{schuster2019towards}             & 90.31\dag       & 78.59       & 70.45     & 69.37     & 70.02\dag     & 75.07   & 55.10\dag      & 32.25\dag & 56.23 \\
+POE~\cite{clark2019don}                  & 90.24\dag       & 78.24       & 69.39     & 68.77     & 68.87     & 75.04   & 55.21\dag      & 32.18\dag & 55.47 \\
+Conf-Reg~\cite{utama2020mind}             & 90.34\dag       & 79.97\dag       & 73.08\dag     & 72.97\dag     & 71.24\dag     & 75.41\dag   & 57.40\dag      & 32.35\dag & 57.27\dag \\
+LIREx~\cite{zhao2021lirex}                & 90.79\textsuperscript{*}       & 79.39\textsuperscript{*}       & -        & 71.55\textsuperscript{*}     & -        & -      & -         & -    & -    \\
+Human Teacher~\cite{pruthi2022evaluating}        & 89.99\textsuperscript{*}       & 79.90\textsuperscript{*}       & -        & 73.27\textsuperscript{*}     & -        & -      & -         & -    & -    \\
+Additional Attention~\cite{stacey2022supervising} & 90.09\textsuperscript{*}       & 79.96\textsuperscript{*}       & 73.10\textsuperscript{*}     & 73.03\textsuperscript{*}     & 71.39\dag     & 75.17   & 57.24\dag      & 31.47\textsuperscript{*} & 57.85\textsuperscript{*} \\
+Existing Attention~\cite{stacey2022supervising}   & 90.17\textsuperscript{*}       & 80.15\textsuperscript{*}       & \textbf{73.19}\textsuperscript{*}     & \textbf{73.36}\textsuperscript{*}     & 71.45\dag     & 75.04   & 57.49\dag      & 31.41\textsuperscript{*} & 58.42\textsuperscript{*} \\
+DC-Match~\cite{zou2022divide}             & 90.25\dag       & 80.04\dag       & 73.18\dag     & 73.19\dag     & 71.33\dag     & 75.23   & 57.12\dag      & 32.21\dag & 58.09\dag \\ \hline
+ATA                  & 90.17       & 79.61\dag       & 73.02\dag     & 72.75     & 71.47\dag     & 75.25   & 57.29\dag      & 31.96 & 58.00\dag \\
+CLS, EBD-Reg              & 90.64\dag       & 79.54       & 72.76     & 73.05\dag     & 71.67\dag     & 75.96   & 57.55\dag      & 32.20\dag & 57.21\dag \\
+ATA, EBD-Reg             & \textbf{91.07}\dag       & \textbf{80.41}\dag       & 72.89\dag     & 73.25\dag     & \textbf{72.07}\dag & \textbf{76.33}\dag     & \textbf{57.74}\dag      & \textbf{32.57}\dag & \textbf{58.96}\dag \\ \hline
\end{tabular}
\caption{The NLI benchmark evaluation results (average accuracy derived from 8 different seeds) of BERT fine-tuned on eSNLI. Notably, \textsuperscript{*} indicates results from the original paper, while \dag signifies a significant improvement in its outcomes (Wilcoxon signed-rank test, $p<0.05$).}~\label{tab.main1}
\end{table*}

\subsubsection{Decoupling}


The essence of EBD-Reg lies in enhancing the model's focus on keywords while discarding its attention to biases. We supervise \textbf{S}elf-\textbf{A}ttention within the $h^{th}$ Transformer block ($\mathcal{L}^h_{\rm SA}$). By supervising the attention weights of the $h^{th}$ Transformer block's $\text{[CLS]}$ token and other tokens, we compel the model to focus more on keywords in the corpus, thereby reducing its learning of biases~\cite{pruthi2022evaluating,stacey2022supervising}. For the input $\bm{s}^{a,b}$ fed into the Transformer-based encoder, the global semantic representation $\{\bm{v}_{h_i}\}_{i=1}^{l_{a,b}}$ can be obtained from each Transformer block, where $\bm{v}_{h_1}$ represents the representation of the first token $\text{[CLS]}$ at the $h^{th}$ Transformer block. The loss for supervising self-attention is formulated in Equation~\ref{eq.lsa}.

\begin{equation}
    \mathcal{L}^h_{\rm SA}=\sum_{i=1}^{l_{a,b}}||\widetilde{e}_i-{\rm attn}^h_i||_2\label{eq.lsa}
\end{equation}

Wherein, we align the desired attention value distribution $\{\widetilde{e}_i\}_{i=1}^{l_{a,b}}$ (normalizing the $e_i$ values) with the normalized weight of each token ${\rm attn}^h_i$ using the $L_2$ norm, as formulated in Equation~\ref{eq.att}.

\begin{equation}
\begin{gathered}
    \widetilde{e}_i=\frac{e_i}{\sum_{k=1}^{l_{a,b}}e_k},\quad
    {\rm attn}^h_i={\rm softmax}(\frac{\bm{v}_{h_{1}}^T \cdot \bm{v}_{h_i}}{\sqrt{l_{a,b}}})\label{eq.att}    
\end{gathered}
\end{equation}

\subsubsection{Aligning}




To highlight the actual contribution of keywords and biases to the final prediction and make more interpretable prediction, we adopt a divide-and-conquer strategy to supervise for \textbf{S}ub-\textbf{I}nferences within the $h^{th}$ Transformer block ($\mathcal{L}^h_{\rm SI}$). We dissect the main inference into two sub-inferences: keyword inference and bias inference, -ressing each separately. Specifically, by masking biases and keywords, we generate two text sequences, $\bm{s}^\psi$ and $\bm{s}^\sigma$. We model the prediction probability distributions for these sub-inferences at the $h^{th}$ Transformer block, denoted as $P^h(y|\bm{s}^{\psi})$ and $P^h(y|\bm{s}^{\sigma})$. Similar to the main inference, we incorporate Token-level Attention to compute these probabilities. 

{
\small
\begin{equation}
\begin{aligned}
P^h(y_\psi|\bm{s}^{\psi})&={\rm softmax}({\bm{W}
}^{\text{pre}}\cdot(\frac{1}{l_{a,b}}\sum_{i=1}^{l_{a,b}}\widetilde{\bm{\lambda}}_{h_i}^{\psi}\bm{v}^{\psi}_{h_i})) \\
P^h(y_\sigma|\bm{s}^{\sigma})&={\rm softmax}({\bm{W}
}^{\text{pre}}\cdot(\frac{1}{l_{a,b}}\sum_{j=1}^{l_{a,b}}\widetilde{\bm{\lambda}}_{h_j}^{\sigma}\bm{v}^{\sigma}_{h_j}))
\end{aligned}
\end{equation}
}

The alignment of the joint probability distribution of sub-inferences with the main inference probability distribution is a cornerstone of interpretability. To compute the joint probability distribution, we assume that each sub-inference should follow the same type of objective and be independent of each other. Based on the inference priority of labels (entailed $>$ neutral $>$ contradicted), we derive the joint probability distribution $P(y|\bm{s}^{\psi},\bm{s}^{\sigma})$, as formulated in Equation~\ref{eq.joint}. Here, $y_1$ and $y_2$ represent target categories reflecting degrees of inference, with $\tau({y_1}) > \tau({y_2})$ indicating that the inference priority of $y_1$ is higher than that of $y_2$. Specifically, for $y\in\{\text{entailed}, \text{neutral}, \text{contradicted}\}$, the corresponding $\tau({y})\in\{2,1,0\}$. The probability $P(y = \text{contradicted}|\bm{s}^{\psi},\bm{s}^\sigma)$ indicates the likelihood of at least one sub-inference being inferred as contradicted, whereas $P(y = \text{entailed}|\bm{s}^{\psi},\bm{s}^\sigma)$ represents the probability of both sub-inferences being inferred as entailed.

\begin{equation}
\begin{gathered}
 \begin{aligned}    &P^h(y=y_1|\bm{s}^{\psi},\bm{s}^\sigma)\\&=P^h(y_\psi=y_1,y_\sigma=y_1|\bm{s}^{\psi},\bm{s}^{\sigma})\\&+\sum_{\tau(y_2)>\tau(y_1)}P^h(y_\psi=y_1,y_\sigma=y_2|\bm{s}^
\psi,\bm{s}^\sigma)\\&+\sum_{\tau(y_2)>\tau(y_1)}P^h(y_\psi=y_2,y_\sigma=y_1|\bm{s}^\psi,\bm{s}^\sigma)
\end{aligned}\\
 P^h(y_{\psi},y_{\sigma}|\bm{s}^{\psi},\bm{s}^{\sigma})=P^h(y_\psi|\bm{s}^{\psi})P^h(y_\sigma|\bm{s}^{\sigma}) 
\end{gathered}~\label{eq.joint}
\end{equation}

Finally, we utilize the JS-divergence to align the main inference probability distribution with the joint probability distribution of sub-inferences, as formulated in Equation~\ref{eq.lsi}. 

{
\small
\begin{equation}
\begin{aligned}
&\mathcal{L}^h_{\rm SI} =\\ 
&\frac{1}{2} (\sum_y P^h(y|\bm{s}^{a,b}) \log \frac{2P^h(y|\bm{s}^{a,b})}{P^h(y|\bm{s}^{a,b}) + P^h(y|\bm{s}^{\psi},\bm{s}^{\sigma})}+ \\ 
 & \sum_y P^h(y|\bm{s}^{\psi},\bm{s}^{\sigma}) \log \frac{2P^h(y|\bm{s}^{\psi},\bm{s}^{\sigma})}{P^h(y|\bm{s}^{a,b}) 
+ P^h(y|\bm{s}^{\psi},\bm{s}^{\sigma})})~\label{eq.lsi}
\end{aligned}
\end{equation}
}

\begin{table*}[t]
\centering
\setlength{\tabcolsep}{4pt} 
\renewcommand{\arraystretch}{1} 
\begin{tabular}{l|ccc|cccccccc}
\hline
\multirow{2}{*}{BERT}         & \multicolumn{3}{c|}{Fine-tune Data}                               & \multicolumn{8}{c}{Evaluated Data}                                                                                                                                                                                                                        \\ \cline{2-12} 
                              & \multicolumn{1}{l}{eSNLI} & \multicolumn{1}{l}{RTE}   & SciTail & \multicolumn{1}{l}{$\rm SNLI_{dev}$} & \multicolumn{1}{l}{$\rm SNLI_{hard}$} & \multicolumn{1}{l}{$\rm MNLI_{m}$} & \multicolumn{1}{l}{$\rm MNLI_{mm}$} & \multicolumn{1}{l}{$\rm QNLI_{dev}$} & \multicolumn{1}{l}{ANLI}  & \multicolumn{1}{l}{HANS}  & Avg.                   \\ \hline
\multirow{3}{*}{CLS}          & \multicolumn{1}{c}{\checkmark} & \multicolumn{1}{l}{}      &         & \multicolumn{1}{l}{90.08}       & \multicolumn{1}{l}{79.36}       & \multicolumn{1}{l}{72.51}     & \multicolumn{1}{l}{72.41}     & \multicolumn{1}{l}{54.89}      & \multicolumn{1}{l}{31.72} & \multicolumn{1}{l}{56.83} & \multirow{3}{*}{66.99} \\ 
                              & \multicolumn{1}{c}{\checkmark} & \multicolumn{1}{c}{\checkmark} &         & \multicolumn{1}{l}{90.23}       & \multicolumn{1}{l}{78.44}       & \multicolumn{1}{l}{72.98}     & \multicolumn{1}{l}{72.01}     & \multicolumn{1}{l}{55.54}      & \multicolumn{1}{l}{32.23} & \multicolumn{1}{l}{55.91} &                        \\ 
                              & \multicolumn{1}{c}{\checkmark} & \multicolumn{1}{c}{\checkmark} & \checkmark   & \multicolumn{1}{l}{90.90}       & \multicolumn{1}{l}{79.43}       & \multicolumn{1}{l}{72.78}     & \multicolumn{1}{l}{72.49}     & \multicolumn{1}{l}{55.79}      & \multicolumn{1}{l}{32.07} & \multicolumn{1}{l}{57.14} &                        \\ \hline
\multirow{3}{*}{ATA}          & \multicolumn{1}{c}{\checkmark} & \multicolumn{1}{l}{}      &         & \multicolumn{1}{l}{90.17}       & \multicolumn{1}{l}{79.61}       & \multicolumn{1}{l}{73.02}     & \multicolumn{1}{l}{72.75}     & \multicolumn{1}{l}{57.29}      & \multicolumn{1}{l}{31.96} & \multicolumn{1}{l}{57.00} & \multirow{3}{*}{67.44} \\ 
                              & \multicolumn{1}{c}{\checkmark} & \multicolumn{1}{c}{\checkmark} &         & \multicolumn{1}{l}{90.82}       & \multicolumn{1}{l}{79.88}       & \multicolumn{1}{l}{72.12}     & \multicolumn{1}{l}{72.34}     & \multicolumn{1}{l}{56.28}      & \multicolumn{1}{l}{32.10} & \multicolumn{1}{l}{55.72} &                        \\ 
                              & \multicolumn{1}{c}{\checkmark} & \multicolumn{1}{c}{\checkmark} & \checkmark   & \multicolumn{1}{l}{90.31}       & \multicolumn{1}{l}{79.94}       & \multicolumn{1}{l}{72.83}     & \multicolumn{1}{l}{72.59}     & \multicolumn{1}{l}{57.67}      & \multicolumn{1}{l}{32.29} & \multicolumn{1}{l}{58.35} &                        \\ \hline
\multirow{3}{*}{ATA, EBD-Reg} & \multicolumn{1}{c}{\checkmark} & \multicolumn{1}{l}{}      &         & \multicolumn{1}{l}{91.07}       & \multicolumn{1}{l}{80.21}       & \multicolumn{1}{l}{72.89}     & \multicolumn{1}{l}{73.25}     & \multicolumn{1}{l}{57.74}      & \multicolumn{1}{l}{32.57} & \multicolumn{1}{l}{58.96} & \multirow{3}{*}{67.90} \\ 
                              & \multicolumn{1}{c}{\checkmark} & \multicolumn{1}{c}{\checkmark} &         & \multicolumn{1}{l}{90.97}       & \multicolumn{1}{l}{79.98}       & \multicolumn{1}{l}{72.76}     & \multicolumn{1}{l}{73.17}     & \multicolumn{1}{l}{57.02}      & \multicolumn{1}{l}{32.47} & \multicolumn{1}{l}{56.81} &                        \\ 
                              & \multicolumn{1}{c}{\checkmark} & \multicolumn{1}{c}{\checkmark} & \checkmark   & \multicolumn{1}{l}{91.45}       & \multicolumn{1}{l}{79.85}       & \multicolumn{1}{l}{72.72}     & \multicolumn{1}{l}{73.78}     & \multicolumn{1}{l}{57.82}      & \multicolumn{1}{l}{32.51} & \multicolumn{1}{l}{57.29} &                        \\ \hline
\end{tabular}
\caption{Introducing further fine-tuning on the SciTail and RTE datasets (train), the evaluation results.}~\label{tab.fur}
\end{table*}


\subsection{Fine-tuning and Evaluation}

During fine-tuning, We construct EBD-Reg as an additional training objective, as described in the bold section of Equation \(\ref{eq.train}\), to aid in the training process. For the hyperparameters \(\alpha\), \(\beta\), and \(H\), we set \(\alpha = 0.4\), \(\beta = 0.8\), and \(H = 3\), signifying that we supervise the top three blocks of the Transformer-based encoder.

\begin{equation}            
\mathcal{L}=\mathcal{L}_{\rm main}+\bm{\alpha\mathcal{L}_{\rm ER}+\frac{\beta}{H}\sum_{h=1}^H(\mathcal{L}^h_{\rm SA}+ \mathcal{L}^h_{\rm SI})}~\label{eq.train}
\end{equation}

During evaluation, we determine the relationship category for the premise-hypothesis pair by directly computing: $y^* = {\rm argmax}_yP(y|\bm{s}^{a,b})$. This implies that the evaluation procedure remains unchanged compared to the backbones and does not require any additional computations. 
The robustness of our approach is evaluated by assessing its accuracy on the following benchmarks: SNLI~\cite{bowman2015large}, SNLI-hard~\cite{gururangan2018annotation}, MNLI~\cite{williams2018broad}, RTE~\cite{giampiccolo2007third}, QNLI~\cite{wang2018glue}, ANLI~\cite{nie2020adversarial}, SciTail~\cite{khot2018scitail} and HANS~\cite{mccoy2019right}.

\begin{figure}[t]
\centering
\includegraphics[width=1\linewidth]{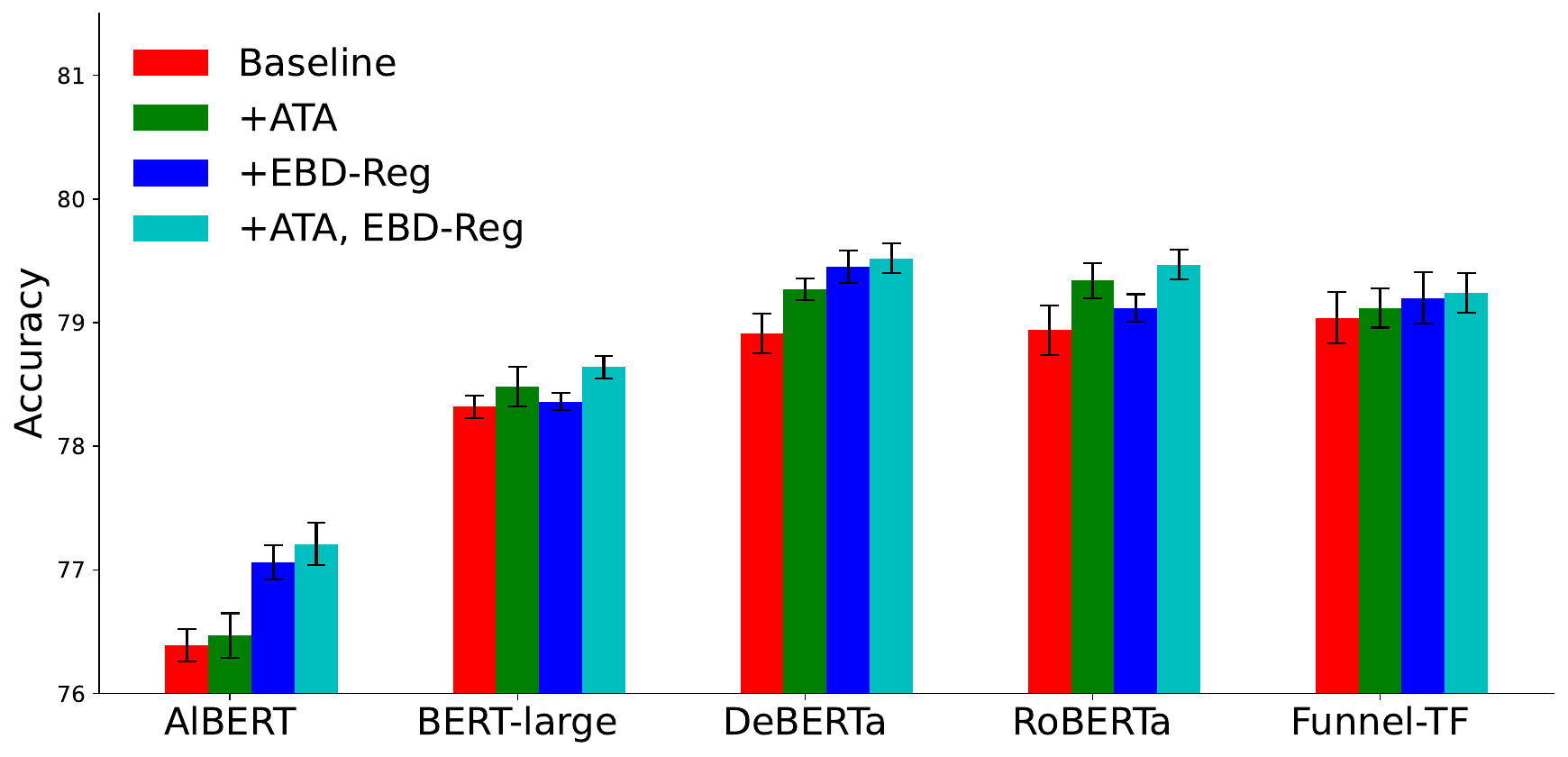}
\caption{
Implementing ATA, EBD-Reg, and their combination on other Transformer-based encoders, where the accuracy is the average performance of SNLI (test), MNLI (mm), RTE (dev). The results are the average of eight random seeds, with the error represented as the standard deviation.}\label{fig.plms}
\end{figure}

\section{Experimental Results and Analysis}

\subsection{Main Results}


Following previous work~\cite{stacey2022supervising}, we compare the impact of introducing various debiasing methods on BERT fine-tuned on eSNLI. The methods includes Reweight~\cite{schuster2019towards}, POE~\cite{clark2019don}, Conf-Reg~\cite{utama2020mind}, LIREx~\cite{zhao2021lirex}, Human Teacher~\cite{pruthi2022evaluating}, supervision on Additional Attention, Existing Attention~\cite{stacey2022supervising}, DC-Match~\cite{zou2022divide} and our EBD-Reg.  
Table~\ref{tab.main1} reports the evaluation accuracy of fine-tuned BERT on NLI benchmarks. The introduction of ATA and EBD-Reg has demonstrated superior generalization performance compared to other methods across multiple benchmarks. Additionally, we investigate the individual impact of introducing ATA and EBD-Reg on BERT. It's observed that introducing ATA led to a more significant improvement in encoder optimization across various datasets compared to BERT alone. When EBD-Reg is introduced separately, the encoder's performance shows even more significant improvement, surpassing the results achieved by using ATA alone, all without the need for additional parameters and inference overhead. Table~\ref{tab.fur} reports the evaluation accuracy on other benchmarks following further fine-tuning on the SciTail and RTE datasets (train) using only $\mathcal{L}_{\rm main}$ for fine-tuning. The combination of ATA and EBD-Reg still maintains the best average performance (67.90). In summary, whether used alone or in combination with ATA, EBD-Reg demonstrates remarkable capabilities, significantly enhancing model performance on out-of-distribution benchmarks. Moreover, this superior performance is sustainable even after the model absorbs other knowledge.

\subsection{Integration with other Encoders}

To further investigate the effectiveness of our proposed training strategy across other Transformer-based encoders, we apply ATA, EBD-Reg, and their combination to AlBERT~\cite{lan2019albert}, BERT-large~\cite{devlin2019bert}, RoBERTa~\cite{liu2019roberta}, DeBERTa~\cite{he2020deberta}, Funnel-TF~\cite{dai2020funnel}, and Figure~\ref{fig.plms} reports the performance changes. Notably, the listed Transformer-based encoders usually possess diverse architectures and parameter scales. We fine-tune each original version of Transformer-based encoders and its variants that incorporated our methods using the same set of configurations, without necessitating additional hyperparameter tuning. We find that the performance improvement manifested by the other Transformer-based encoders after the introduction of our methods is highly consistent with the results on BERT. This indicates the strong universality and stability of EBD-Reg across dataset and backbone selections.

\subsection{Hyperparameters $\alpha$,$\beta$,$H$}

\begin{figure}[t]
\centering
\includegraphics[width=1\linewidth]{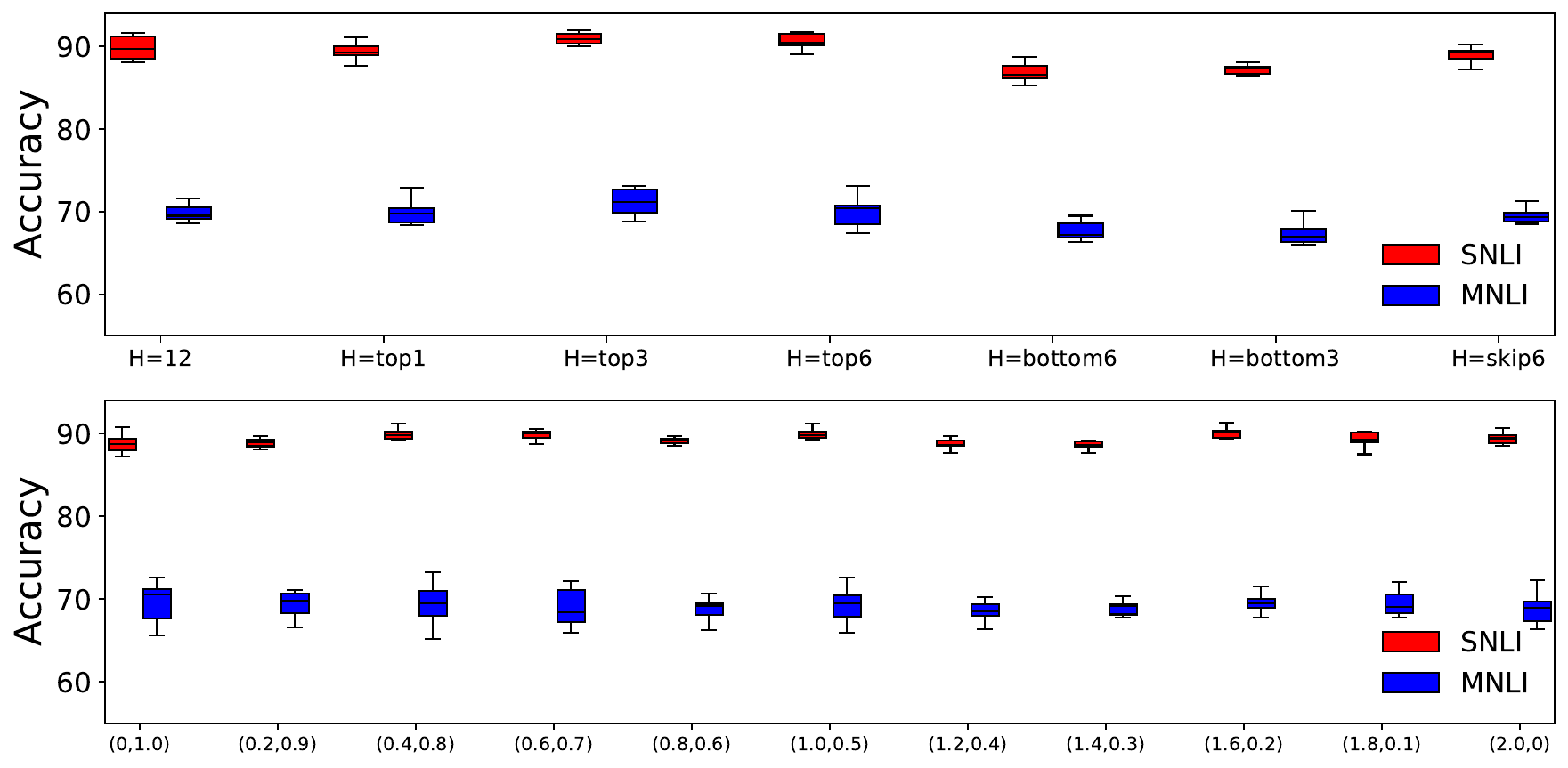}
\caption{The effect of \( \alpha \),\( \beta \) ,\( H \)}\label{fig.box}
\end{figure}




We further discuss the impact of hyperparameters \( \alpha \), \( \beta \), and \( H \) using BERT as the backbone. In our discussion of the influence of hyperparameter \( H \), we devised seven supervision strategies: supervising all 12 Transformer blocks, only the topmost Transformer block, the top 3 Transformer blocks, the top 6 Transformer blocks, the bottom 6 Transformer blocks, the bottom 3 Transformer blocks, and 6 alternating Transformer blocks. For BERT, we tested these strategies using eight different combinations of \( \alpha \) and \( \beta \) hyperparameters. Figure~\ref{fig.box} presents the distribution of results on SNLI (dev) and MNLI (mismatched).

For the evaluation, strategies using all 12 Transformer blocks, and the top 1, 3, and 6 blocks, on average, performed better than those using the bottom 6, 3, and alternating 6 blocks. This can be attributed to the deeper Transformer blocks providing more refined semantic information after multiple layers of self-attention and feedforward networks. Among these top block supervision strategies, the top 3 blocks demonstrated excellent stability and robustness in in-distribution data while ensuring high precision. 

In discussing the impact of hyperparameters \( \alpha \) and \( \beta \), we advocate that in the training objective, the loss from EBD-Reg should be double that of the main inference loss, i.e., \( \alpha+2\beta=2 \). Hence, we conducted experiments with 11 sets of hyperparameter combinations, each tested against the seven different \( H \) selections. Figure~\ref{fig.box} also shows the distribution of results for these 11 sets on eSNLI (dev) and MNLI (mismatched). In SNLI inference, the hyperparameter choice of \( (\alpha, \beta)=(0.4,0.8) \) ensured model robustness while maintaining inference precision. This indicates that the importance of Decoupling and Aligning in in-distribution data inference outweighs the identification of keywords and biases. 

\begin{table}[t]
\centering
\footnotesize
\setlength{\tabcolsep}{7pt} 
\renewcommand{\arraystretch}{1} 
\begin{tabular}{lll}
\hline
Model           & Accuracy ($\uparrow$)      & Token-level Score ($\uparrow$) \\ \hline
BERT+$\mathcal{L}_{\rm main}$            & 78.03         & 26.85             \\
+$\mathcal{L}_{\rm ER}$         & 77.53 (-0.50) & 62.45 (35.60)     \\
+$\mathcal{L}^h_{\rm SA}$         & 78.16 (0.17)  & 41.45 (14.60)     \\
+$\mathcal{L}^h_{\rm SI}$         & 78.29 (0.30)  & 21.56 (-5.29)     \\
+$\mathcal{L}_{\rm ER}$, $\mathcal{L}^h_{\rm SA}$     & 78.67 (0.68)  & 63.89 (37.04)     \\
+$\mathcal{L}_{\rm ER}$, $\mathcal{L}^h_{\rm SI}$     & 78.55 (0.56)  & 64.21 (37.36)     \\
+$\mathcal{L}^h_{\rm SI}$, $\mathcal{L}^h_{\rm SA}$     & 78.64 (0.65)  & 28.77 (1.92)      \\
+$\mathcal{L}_{\rm ER}$, $\mathcal{L}^h_{\rm SA}$, $\mathcal{L}^h_{\rm SI}$ & 78.80 (0.81)   & 64.10 (37.25)     \\ \hline
DeBERTa+$\mathcal{L}_{\rm main}$         & 79.19         & 50.06             \\
+$\mathcal{L}_{\rm ER}$         & 79.08 (-0.11) & 67.75 (17.69)     \\
+$\mathcal{L}^h_{\rm SA}$         & 79.39 (0.20)  & 59.56 (9.50)      \\
+$\mathcal{L}^h_{\rm SI}$         & 79.25 (0.06)  & 48.18 (-1.88)     \\
+$\mathcal{L}_{\rm ER}$, $\mathcal{L}^h_{\rm SA}$     & 79.62 (0.44)  & 69.24 (19.17)     \\
+$\mathcal{L}_{\rm ER}$, $\mathcal{L}^h_{\rm SI}$     & 79.73 (0.54)  & 69.14 (19.08)     \\
+$\mathcal{L}^h_{\rm SI}$, $\mathcal{L}^h_{\rm SA}$     & 79.74 (0.55)  & 59.46 (9.40)      \\
+$\mathcal{L}_{\rm ER}$, $\mathcal{L}^h_{\rm SA}$, $\mathcal{L}^h_{\rm SI}$ & 79.86 (0.67)  & 69.21 (19.15) \\ \hline   
\end{tabular}
\caption{Ablation study to validate the effectiveness of each supervision target in EBD-Reg. We report the average accuracies on the evaluations of SNLI (test), MNLI (mm), and RTE (dev), token-level scores (F1), and their improvements.}~\label{tab.albtion}
\end{table}

\begin{figure}[t]
\centering
\includegraphics[width=1\linewidth]{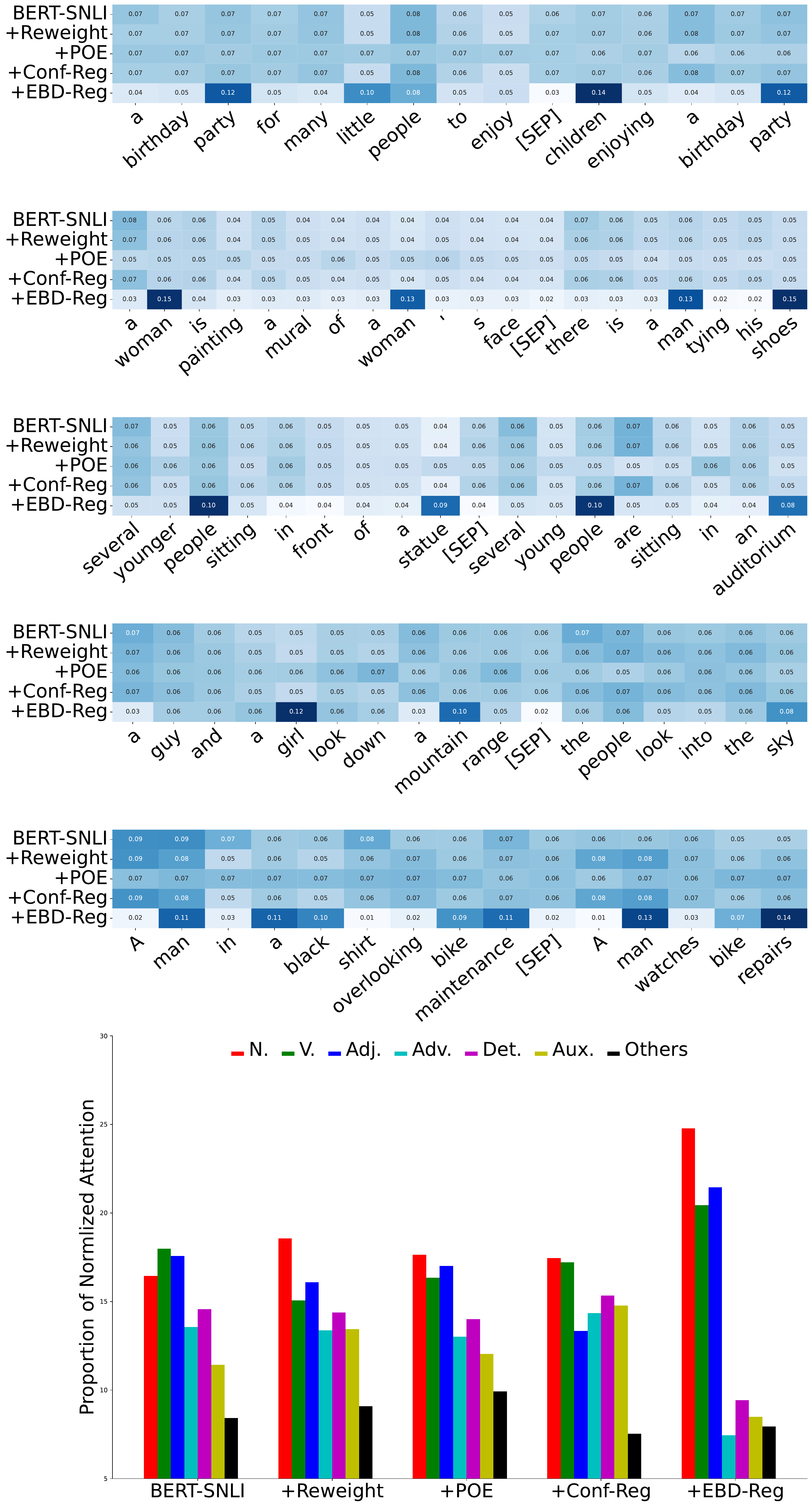}
\caption{
Case study concerning supervised for self-attention. The heatmap reports the normalized attention values of the $\text{[CLS]}$ token and other tokens in sentences. The bar chart presents the average proportions of normalized attention allocated to each part of speech.}\label{fig.heapmap}
\end{figure}

\begin{table*}[t]
\centering
\renewcommand{\arraystretch}{1} 
\begin{tabular}{lccccc}
\hline
Premise ($\bm{s}^a$) \& Hypothesis ($\bm{s}^b$)                                               & Label $y$            & BERT               & +EBD-Reg & Keywords           & Biases               \\\hline
$\bm{s}^a$:A birthday party for many \hl{little people} to enjoy.      & \multirow{2}{*}{N} & \multirow{2}{*}{C} & \multirow{2}{*}{N}   & \multirow{2}{*}{N} & \multirow{2}{*}{E} \\
$\bm{s}^b$:\hl{Children} enjoying a birthday party.                    &                    &                    &                      &                    &                    \\\hline
$\bm{s}^a$:\hl{A woman} is painting a mural of a woman's face.         & \multirow{2}{*}{C} & \multirow{2}{*}{E} & \multirow{2}{*}{C}   & \multirow{2}{*}{E} & \multirow{2}{*}{C} \\
$\bm{s}^b$:There is \hl{a man} tying his shoes.                         &                    &                    &                      &                    &                    \\\hline
$\bm{s}^a$:Several younger \hl{people sitting} in front of \hl{a statue}. & \multirow{2}{*}{C} & \multirow{2}{*}{N} & \multirow{2}{*}{C}   & \multirow{2}{*}{C} & \multirow{2}{*}{N} \\
$\bm{s}^b$:Several young \hl{people} are \hl{sitting} in an \hl{auditorium}.  &                    &                    &                      &                    &                    \\\hline
$\bm{s}^a$:A guy and a girl \hl{look down} a mountain range.           & \multirow{2}{*}{C} & \multirow{2}{*}{C} & \multirow{2}{*}{C}   & \multirow{2}{*}{C} & \multirow{2}{*}{C} \\
$\bm{s}^b$:The people \hl{look into the sky}.                          &                    &                    &                      &                    &  \\ \hline
$\bm{s}^a$:A man in a black shirt overlooking \hl{bike maintenance}.   & \multirow{2}{*}{E} & \multirow{2}{*}{E} & \multirow{2}{*}{E}   & \multirow{2}{*}{E} & \multirow{2}{*}{E} \\
$\bm{s}^b$:A man watches \hl{bike repairs}.                            &                    &                    &                      &                    &                    \\\hline
\end{tabular}
\caption{Case study concerning supervised for sub-inferences, where highlighted sections denote keywords and the remaining parts represent biases labeled by explanations. "Label", "BERT", "+EBD-Reg", "Keywords", and "Biases" correspond to: Label, predictions made with BERT as the backbone, predictions made after integrating EBD-Reg, and predictions of sub-inferences (keyword inference, bias inference).}~\label{tab.case}
\end{table*}

\subsection{Ablation Study}

We also conduct ablation studies to validate the effectiveness of each supervision target within EBD-Reg. Table~\ref{tab.albtion} reports the inference accuracy (Accuracy) and Token-level Score (F1) as evaluation metrics. For both BERT and DeBERTa, when the three training targets $\mathcal{L}_{\rm ER}$, $\mathcal{L}^h_{\rm SA}$, and $\mathcal{L}^h_{\rm SI}$ are introduced simultaneously, the encoder's performance achieves its maximum improvement. Among all the training targets introduced individually, sub-inference supervision ($\mathcal{L}^h_{\rm SI}$) offers the most significant performance boost for the BERT, while the sub self-attention ($\mathcal{L}^h_{\rm SA}$) results in the largest improvement for DeBERTa. If using  entity recognition supervision ($\mathcal{L}_{\rm ER}$) solely, although we observed a significant increase in the encoder's ability to distinguish between keywords and biases (Token-level Score), there was no notable improvement in inference accuracy compared to the backbones. This suggests that the distinguishing between keywords and biases might not be directly correlated with the main inference. When multiple training targets are introduced simultaneously, the combination of $\mathcal{L}_{\rm ER}$ and $\mathcal{L}^h_{\rm SA}$ yields the greatest performance enhancement for the BERT, while the combination of $\mathcal{L}_{\rm ER}$ and $\mathcal{L}^h_{\rm SI}$ does the same for the DeBERTa. This indicates that distinguishing between keywords and biases assists the encoder in obtaining content representations that require varying levels of inference granularity. For both backbones, introducing a combination of three training objectives resulted in the greatest performance enhancement. This suggests that these three training objectives can complement each other, collectively optimizing the model's performance.


\subsection{Case Study}
To intuitively investigate the workings of EBD-Reg, we have selected several premise-hypothesis pairs from the SNLI test set for a case study focusing on the supervision of self-attention and sub-inferences within EBD-Reg.

\subsubsection{Attention Analysis}

We independently obtained the encoded sentence pairs from the fine-tuned BERT model and the models modified with four different debiasing methods (+Reweight, +POE, +Conf-Reg, +EBD-Reg). We then gathered the attention scores of the \text{[CLS]} token representations and other word representations. These scores reflect the degree of influence each word in the sentence has on the overall semantic meaning. Notably, the baseline model and the models incorporating Reweight, POE, and Conf-Reg demonstrated similar attention distribution to varying content during inference. In contrast, the introduction of the EBD-Reg model identified keywords in the premises and hypotheses. For instance, as exemplified in sentence four in Figure~\ref{fig.heapmap}, the analysis of the model’s attention behavior post the introduction of EBD-Reg revealed a shift in focus. After introducing EBD-Reg, unlike the baseline that gives almost uniform attention to all words, BERT now directs its attention to "mountain" in the premise and "sky" in the hypothesis. These concrete nouns are crucial for identifying the contradictory relationship between the premise and hypothesis, as elucidated by the corresponding explanation "One can't look down into the sky". Furthermore, we utilized NLTK~\footnote{https://www.nltk.org/} to perform word-level part-of-speech tagging on all samples in the SNLI test set, and calculated the proportion of normalized attention scores for each part of speech category (further standardized by the number of words). From a part-of-speech perspective, with the introduction of EBD-Reg, the model is able to allocate more attention to nouns, verbs, adjectives, and other words representing entities, actions, events, etc., which are more likely to be relevant to the intended task. In contrast, the baseline and other methods show less variability in the distribution of attention across different parts of speech.


\subsubsection{Consistency Analysis}
We directly implement predictions on two sub-inferences: $y^\psi = {\rm argmax}_yP^h(y|\bm{s}^{\psi}), h=H$ and $y^\sigma = {\rm argmax}_yP^h(y|\bm{s}^{\sigma}), h=H$. As demonstrated in Table~\ref{tab.case}, we observe that after incorporating EBD-Reg, there is a significant alignment between BERT's global inference predictions and the ultimate predictions of the sub-inferences. Acquiring higher inference priority predictions in both sub-inferences simultaneously is a prerequisite for the global prediction to obtain higher inference results. On the other hand, if any sub-argument is inferred to yield a lower inference priority, it becomes a sufficient condition for the global inference to predict an outcome with a lower priority (the inference priorities are, in descending order: Entailed (E), Neutral (N), Contradicted (C)).

\section{Conclusion}


In conclusion, while traditional NLI debiasing methods teach models to learn "which samples are biased", we aim to instruct models on "which parts of a sample are biased". Inspired by the logic in human explanations of causality, we conducted an initial analysis of the intrinsic connection between human explanations and biases, proposing a standard for dividing keywords and biases, and a simple, comprehensive, and interpretable debiasing method: Explanation based Bias Decoupling Regularization (EBD-Reg). EBD-Reg can be easily integrated with various Transformer-based encoders, significantly outperforming other methods in out-of-distribution inference.



%



\bibliography{ecai}

\begin{thebibliography}{10}
\providecommand{\url}[1]{#1}
\csname url@samestyle\endcsname
\providecommand{\newblock}{\relax}
\providecommand{\bibinfo}[2]{#2}
\providecommand{\BIBentrySTDinterwordspacing}{\spaceskip=0pt\relax}
\providecommand{\BIBentryALTinterwordstretchfactor}{4}
\providecommand{\BIBentryALTinterwordspacing}{\spaceskip=\fontdimen2\font plus
\BIBentryALTinterwordstretchfactor\fontdimen3\font minus \fontdimen4\font\relax}
\providecommand{\BIBforeignlanguage}[2]{{%
\expandafter\ifx\csname l@#1\endcsname\relax
\typeout{** WARNING: IEEEtran.bst: No hyphenation pattern has been}%
\typeout{** loaded for the language `#1'. Using the pattern for}%
\typeout{** the default language instead.}%
\else
\language=\csname l@#1\endcsname
\fi
#2}}
\providecommand{\BIBdecl}{\relax}
\BIBdecl

\bibitem{devlin2019bert}
J.~Devlin, M.-W. Chang, K.~Lee, and K.~Toutanova, ``Bert: Pre-training of deep bidirectional transformers for language understanding,'' in \emph{Proceedings of the 2019 Conference of the North American Chapter of the Association for Computational Linguistics: Human Language Technologies, Volume 1 (Long and Short Papers)}, 2019, pp. 4171--4186.

\bibitem{liu2019roberta}
Y.~Liu, M.~Ott, N.~Goyal, J.~Du, M.~Joshi, D.~Chen, O.~Levy, M.~Lewis, L.~Zettlemoyer, and V.~Stoyanov, ``Roberta: A robustly optimized bert pretraining approach,'' \emph{arXiv preprint arXiv:1907.11692}, 2019.

\bibitem{lan2019albert}
Z.~Lan, M.~Chen, S.~Goodman, K.~Gimpel, P.~Sharma, and R.~Soricut, ``Albert: Alite bert for self-supervised learning of language representations,'' \emph{arXiv preprint arXiv:1909.11942}, 2019.

\bibitem{zang2023improving}
J.~Zang and H.~Liu, ``Improving text semantic similarity modeling through a 3d siamese network,'' in \emph{ECAI 2023}.\hskip 1em plus 0.5em minus 0.4em\relax IOS Press, 2023, pp. 2970--2977.

\bibitem{bowman2015large}
S.~Bowman, G.~Angeli, C.~Potts, and C.~D. Manning, ``A large annotated corpus for learning natural language inference,'' in \emph{Proceedings of the 2015 Conference on Empirical Methods in Natural Language Processing}, 2015, pp. 632--642.

\bibitem{gururangan2018annotation}
S.~Gururangan, S.~Swayamdipta, O.~Levy, R.~Schwartz, S.~Bowman, and N.~A. Smith, ``Annotation artifacts in natural language inference data,'' in \emph{Proceedings of the 2018 Conference of the North American Chapter of the Association for Computational Linguistics: Human Language Technologies, Volume 2 (Short Papers)}, 2018, pp. 107--112.

\bibitem{mccoy2019right}
T.~McCoy, E.~Pavlick, and T.~Linzen, ``Right for the wrong reasons: Diagnosing syntactic heuristics in natural language inference,'' in \emph{Proceedings of the 57th Annual Meeting of the Association for Computational Linguistics}, 2019, pp. 3428--3448.

\bibitem{schuster2019towards}
T.~Schuster, D.~Shah, Y.~J.~S. Yeo, D.~R.~F. Ortiz, E.~Santus, and R.~Barzilay, ``Towards debiasing fact verification models,'' in \emph{Proceedings of the 2019 Conference on Empirical Methods in Natural Language Processing and the 9th International Joint Conference on Natural Language Processing (EMNLP-IJCNLP)}, 2019, pp. 3419--3425.

\bibitem{clark2019don}
C.~Clark, M.~Yatskar, and L.~Zettlemoyer, ``Don’t take the easy way out: Ensemble based methods for avoiding known dataset biases,'' in \emph{Proceedings of the 2019 Conference on Empirical Methods in Natural Language Processing and the 9th International Joint Conference on Natural Language Processing (EMNLP-IJCNLP)}, 2019, pp. 4069--4082.

\bibitem{utama2020mind}
P.~A. Utama, N.~S. Moosavi, and I.~Gurevych, ``Mind the trade-off: Debiasing nlu models without degrading the in-distribution performance,'' in \emph{Proceedings of the 58th Annual Meeting of the Association for Computational Linguistics}, 2020, pp. 8717--8729.

\bibitem{camburu2018snli}
O.-M. Camburu, T.~Rockt{\"a}schel, T.~Lukasiewicz, and P.~Blunsom, ``e-snli: Natural language inference with natural language explanations,'' \emph{Advances in Neural Information Processing Systems}, vol.~31, 2018.

\bibitem{minervini2018adversarially}
P.~Minervini and S.~Riedel, ``Adversarially regularising neural nli models to integrate logical background knowledge,'' in \emph{Proceedings of the 22nd Conference on Computational Natural Language Learning}, 2018, pp. 65--74.

\bibitem{min2020syntactic}
J.~Min, R.~T. McCoy, D.~Das, E.~Pitler, and T.~Linzen, ``Syntactic data augmentation increases robustness to inference heuristics,'' in \emph{Proceedings of the 58th Annual Meeting of the Association for Computational Linguistics}, 2020, pp. 2339--2352.

\bibitem{teney2020learning}
D.~Teney, E.~Abbasnedjad, and A.~van~den Hengel, ``Learning what makes a difference from counterfactual examples and gradient supervision,'' in \emph{Computer Vision--ECCV 2020: 16th European Conference, Glasgow, UK, August 23--28, 2020, Proceedings, Part X 16}.\hskip 1em plus 0.5em minus 0.4em\relax Springer, 2020, pp. 580--599.

\bibitem{belinkov2020variational}
Y.~Belinkov, J.~Henderson \emph{et~al.}, ``Variational information bottleneck for effective low-resource fine-tuning,'' in \emph{International Conference on Learning Representations}, 2020.

\bibitem{utama2020towards}
P.~A. Utama, N.~S. Moosavi, and I.~Gurevych, ``Towards debiasing nlu models from unknown biases,'' in \emph{Proceedings of the 2020 Conference on Empirical Methods in Natural Language Processing (EMNLP)}, 2020, pp. 7597--7610.

\bibitem{he2019unlearn}
H.~He, S.~Zha, and H.~Wang, ``Unlearn dataset bias in natural language inference by fitting the residual,'' in \emph{Proceedings of the 2nd Workshop on Deep Learning Approaches for Low-Resource NLP (DeepLo 2019)}, 2019, pp. 132--142.

\bibitem{liu2020empirical}
T.~Liu, Z.~Xin, X.~Ding, B.~Chang, and Z.~Sui, ``An empirical study on model-agnostic debiasing strategies for robust natural language inference,'' in \emph{Proceedings of the 24th Conference on Computational Natural Language Learning}, 2020, pp. 596--608.

\bibitem{sanh2020learning}
V.~Sanh, T.~Wolf, Y.~Belinkov, and A.~M. Rush, ``Learning from others' mistakes: Avoiding dataset biases without modeling them,'' \emph{arXiv preprint arXiv:2012.01300}, 2020.

\bibitem{ghaddar2021end}
A.~Ghaddar, P.~Langlais, M.~Rezagholizadeh, and A.~Rashid, ``End-to-end self-debiasing framework for robust nlu training,'' in \emph{Findings of the Association for Computational Linguistics: ACL-IJCNLP 2021}, 2021, pp. 1923--1929.

\bibitem{belinkov2019adversarial}
Y.~Belinkov, A.~Poliak, S.~M. Shieber, B.~Van~Durme, and A.~Rush, ``On adversarial removal of hypothesis-only bias in natural language inference,'' \emph{NAACL HLT 2019}, p. 256, 2019.

\bibitem{belinkov2019don}
Y.~Belinkov, A.~Poliak, S.~M. Shieber, B.~Van~Durme, and A.~M. Rush, ``Don’t take the premise for granted: Mitigating artifacts in natural language inference,'' in \emph{Proceedings of the 57th Annual Meeting of the Association for Computational Linguistics}, 2019, pp. 877--891.

\bibitem{tu2020empirical}
L.~Tu, G.~Lalwani, S.~Gella, and H.~He, ``An empirical study on robustness to spurious correlations using pre-trained language models,'' \emph{Transactions of the Association for Computational Linguistics}, vol.~8, pp. 621--633, 2020.

\bibitem{zou2022divide}
Y.~Zou, H.~Liu, T.~Gui, J.~Wang, Q.~Zhang, M.~Tang, H.~Li, and D.~Wang, ``Divide and conquer: Text semantic matching with disentangled keywords and intents,'' in \emph{Findings of the Association for Computational Linguistics: ACL 2022}, 2022, pp. 3622--3632.

\bibitem{rei2018jointly}
M.~Rei and A.~S{\o}gaard, ``Jointly learning to label sentences and tokens,'' \emph{arXiv preprint arXiv:1811.05949}, 2018.

\bibitem{stacey2022supervising}
J.~Stacey, Y.~Belinkov, and M.~Rei, ``Supervising model attention with human explanations for robust natural language inference,'' in \emph{Proceedings of the AAAI Conference on Artificial Intelligence}, 2022.

\bibitem{bommasani2021opportunities}
R.~Bommasani, D.~A. Hudson, E.~Adeli, R.~Altman, S.~Arora, S.~von Arx, M.~S. Bernstein, J.~Bohg, A.~Bosselut, E.~Brunskill \emph{et~al.}, ``On the opportunities and risks of foundation models,'' \emph{arXiv preprint arXiv:2108.07258}, 2021.

\bibitem{su2021keep}
Y.~Su, D.~Vandyke, S.~Baker, Y.~Wang, and N.~Collier, ``Keep the primary, rewrite the secondary: A two-stage approach for paraphrase generation,'' in \emph{Findings of the Association for Computational Linguistics: ACL-IJCNLP 2021}, 2021, pp. 560--569.

\bibitem{miller1995wordnet}
G.~A. Miller, ``Wordnet: a lexical database for english,'' \emph{Communications of the ACM}, vol.~38, no.~11, pp. 39--41, 1995.

\bibitem{pennington2014glove}
J.~Pennington, R.~Socher, and C.~D. Manning, ``Glove: Global vectors for word representation,'' in \emph{Proceedings of the 2014 Conference on Empirical Methods in Natural lLanguage Processing (EMNLP 2014)}, 2014, pp. 1532--1543.

\bibitem{zhao2021lirex}
X.~Zhao and V.~V. Vydiswaran, ``Lirex: Augmenting language inference with relevant explanations,'' in \emph{Proceedings of the AAAI Conference on Artificial Intelligence}, 2021.

\bibitem{pruthi2022evaluating}
D.~Pruthi, R.~Bansal, B.~Dhingra, L.~B. Soares, M.~Collins, Z.~C. Lipton, G.~Neubig, and W.~W. Cohen, ``Evaluating explanations: How much do explanations from the teacher aid students?'' \emph{Transactions of the Association for Computational Linguistics}, vol.~10, pp. 359--375, 2022.

\bibitem{williams2018broad}
A.~Williams, N.~Nangia, and S.~Bowman, ``A broad-coverage challenge corpus for sentence understanding through inference,'' in \emph{Proceedings of the 2018 Conference of the North American Chapter of the Association for Computational Linguistics: Human Language Technologies, Volume 1 (Long Papers)}, 2018, pp. 1112--1122.

\bibitem{giampiccolo2007third}
D.~Giampiccolo, B.~Magnini, I.~Dagan, and W.~B. Dolan, ``The third pascal recognizing textual entailment challenge,'' in \emph{Proceedings of the ACL-PASCAL workshop on textual entailment and paraphrasing}, 2007, pp. 1--9.

\bibitem{wang2018glue}
A.~Wang, A.~Singh, J.~Michael, F.~Hill, O.~Levy, and S.~Bowman, ``Glue: A multi-task benchmark and analysis platform for natural language understanding,'' in \emph{Proceedings of the 2018 EMNLP Workshop BlackboxNLP: Analyzing and Interpreting Neural Networks for NLP}.\hskip 1em plus 0.5em minus 0.4em\relax Association for Computational Linguistics, 2018.

\bibitem{nie2020adversarial}
Y.~Nie, A.~Williams, E.~Dinan, M.~Bansal, J.~Weston, and D.~Kiela, ``Adversarial nli: A new benchmark for natural language understanding,'' in \emph{Proceedings of the 58th Annual Meeting of the Association for Computational Linguistics}, 2020, pp. 4885--4901.

\bibitem{khot2018scitail}
T.~Khot, A.~Sabharwal, and P.~Clark, ``Scitail: a textual entailment dataset from science question answering,'' in \emph{Proceedings of the Thirty-Second AAAI Conference on Artificial Intelligence and Thirtieth Innovative Applications of Artificial Intelligence Conference and Eighth AAAI Symposium on Educational Advances in Artificial Intelligence}, 2018, pp. 5189--5197.

\bibitem{he2020deberta}
P.~He, X.~Liu, J.~Gao, and W.~Chen, ``Deberta: Decoding-enhanced bert with disentangled attention,'' \emph{arXiv preprint arXiv:2006.03654}, 2020.

\bibitem{dai2020funnel}
Z.~Dai, G.~Lai, Y.~Yang, and Q.~V. Le, ``Funnel-transformer: filtering out sequential redundancy for efficient language processing,'' in \emph{Proceedings of the 34th International Conference on Neural Information Processing Systems}, 2020, pp. 4271--4282.

\end{thebibliography}
\bibliographystyle{IEEEtran}

\end{document}